\documentclass[sigconf]{acmart}

\usepackage{booktabs}

\copyrightyear{2026}
\acmYear{2026}
\setcopyright{cc}
\setcctype{by}
\acmDOI{10.1145/3858063.3858882}
\acmISBN{979-8-4007-3087-0/2026/11}
\acmConference[ICAIF '26]{7th ACM International Conference on AI in Finance}{November 14--17, 2026}{Milan, Italy}
\acmBooktitle{7th ACM International Conference on AI in Finance (ICAIF '26), November 14--17, 2026, Milan, Italy}

\newcommand{\dopfn}{M-DoT-PFN}
\newcommand{\dodo}[1]{\mathrm{do}(#1)}

\begin{document}

\title{Market Impact Estimation via Causal Foundation Models}

\author{Dennis Thumm}
\orcid{0009-0001-3666-2415}
\affiliation{%
  \institution{National University of Singapore}
  \city{Singapore}
  \country{Singapore}}
\email{dennis.thumm@u.nus.edu}

\author{Ying Chen}
\orcid{0000-0002-2577-7348}
\affiliation{%
  \institution{National University of Singapore}
  \city{Singapore}
  \country{Singapore}}
\email{matcheny@nus.edu.sg}

\renewcommand{\shortauthors}{Thumm and Chen}

\begin{abstract}
When a trader executes a large order, the relevant question is not \emph{what will the price be?} but \emph{what would the price have been had I not traded?} Market impact is thus a causal quantity, yet classical models regress price on flow, which is biased whenever informed flow moves prices through information as well as trading. We present a proof-of-concept \emph{Causal Market Simulator}: a continuous-time causal prior-data fitted network (PFN) trained on a synthetic market prior in which a hidden informational flow confounds the flow--price relation, with confounding strength an exact per-episode knob. The naive estimator's bias follows a closed form we confirm empirically (up to 68\% overestimation), while the PFN approaches the unbiased oracle in the high-confounding regime. Its \emph{do-ablated twin}, the same model without its intervention inputs, collapses onto the pre-window mean. Zero-shot on ABIDES, the paired causal-minus-ablated prediction difference is $+0.38$ pre-onset standard deviations, while independently measured seed-replay mean impact is $+0.28$. On about 2,300 detected flow bursts in BTC/ETH/SOL perpetuals and 3,800 Nasdaq equity bursts, the intervention-sensitive predictions are consistent with market impact but cannot establish causal identification without counterfactuals. Off-prior the response degrades gracefully, and split-conformal recalibration restores coverage on held-out ABIDES counterfactuals, while real-market calibration is necessarily factual. Transfer hinges on encoding the metaorder as added flow rather than a clamped flow level. A recurrent state-space variant closes the remaining shape gap: its impact accrues over execution duration where the transformer's stays flat, raising the correlation with measured metaorder impact from $+0.68$ to between $+0.92$ and $+0.96$, though absolute magnitude still needs per-domain calibration and its training recipe weakens on-prior deconfounding.
\end{abstract}



\maketitle

\section{Introduction}\label{sec:intro}

Market impact, the price move caused by one's own trading, is the central cost of execution and the canonical quantity of market microstructure. It is defined as the realized price path \emph{minus the counterfactual path that would have occurred without the trade}~\cite{bonart2026diffusive}. Classical impact models are predictive approximations of this interventional object. Almgren--Chriss fits a linear permanent-impact coefficient by regressing price on flow~\cite{almgren2001optimal}. Propagator models fit a decay kernel to the same observational data~\cite{obizhaeva2013optimal,bouchaud2009markets}. Their observational fits can target $E[Y \mid A]$ rather than the interventional object $E[Y \mid \dodo{A}]$, the price under an \emph{imposed} trade. The two differ whenever something moves both order flow and prices, as informed trading can do in real markets~\cite{muhle2026unified}. Informed traders buy \emph{because} they expect the price to rise, so their flow predicts price moves it did not mechanically cause. Estimated this way, impact coefficients absorb the information channel, and execution-cost models inherit the bias.

Recent foundation models for microstructure learn rich \emph{observational} models of order flow at scale~\cite{sood2026tradefm}, but have no do-operator~\cite{pearl2009causality}: they answer ``what typically happens next,'' not ``what happens if I trade.'' Conversely, recent causal amortized-inference models, prior-data fitted networks (PFNs) trained to perform Bayesian causal inference in context, operate on tabular~\cite{robertson2025dopfn, balazadeh2025causalpfn, dhir2025estimating, ma2026foundation, reuter2026use} or discrete-time~\cite{thumm2026causal} data. The continuous-time construction in~\cite{thumm2026towards} admits irregular event schedules as in markets, but like others has been evaluated primarily on synthetic or simulator-based tasks, the evidence gap our transfer tiers address. A generative line separately argued for causal market simulators~\cite{thumm2025towards}. What is missing is the bridge: a continuous-time \emph{causal} model, specialized to a market prior in which the causal estimand is provably distinct from the predictive one, and evidence that it transfers beyond its prior.

This paper delivers that proof of concept. Our contributions:
\begin{itemize}
  \item \textbf{A market prior with controllable confounding (\S\ref{sec:prior}).} We split order flow into hidden informational \emph{core} flow and endogenous \emph{reaction} flow~\cite{muhle2026unified}, and couple the core into the price through an informational channel $\gamma$. Per-episode confounding strength is an exact, samplable quantity, and the naive estimator's bias is available in closed form. It is linear in $\gamma$ and zero when either channel closes~\cite{pearl2009causality}.
  \item \textbf{Implicit deconfounding, verified against an analytic law (\S\ref{sec:synthetic}).} On episodes from this prior, the naive regression overestimates impact by up to 68\% at strong coupling, matching the closed form within 2\%. The causal PFN tracks the unbiased true-parameter oracle across the entire $\gamma$ range, while its do-ablated twin, the identical model trained and evaluated with its five intervention inputs zeroed (\S\ref{sec:model}), is statistically indistinguishable from the pre-window-mean baseline. The matched ablation attributes the learned intervention response to the do-input pathway.
  \item \textbf{Zero-shot transfer with matched semantics (\S\ref{sec:transfer}).} Without any retraining, the same model responds at the ground-truth scale to metaorders~\cite{maitrier2025generating} in an ABIDES~\cite{byrd2020abides} limit-order book (paired \emph{do-head} effect, causal minus ablated twin, of $+0.38\,\hat{\sigma}$, 95\% CI $[0.34, 0.42]$, versus measured seed-replay mean impact $+0.28\,\hat{\sigma}$, 95\% CI $[0.05, 0.50]$, on 1,000 held-out episodes) and reduces prediction bias around detected flow bursts on BTC/ETH/SOL perpetuals (paired \emph{do-head} effect of $+1.2$--$1.5\,\hat{\sigma}$ over ${\approx}760$ events each; Table~\ref{tab:transfer}). Two mechanism findings enable the transfer: prior-scale standardization and matching the intervention's \emph{semantics} (additive flow, not a clamp) to the event's (\S\ref{sec:abides}).
  \item \textbf{An architecture whose response accrues over execution (\S\ref{sec:model}).} A recurrent state-space variant, gated on inter-observation time, composes impact in its recurrent state, lifting the 12-cell metaorder-benchmark correlation of \S\ref{sec:crypto} from $+0.68$ to $+0.92$--$+0.96$ across training seeds.
\end{itemize}

\section{Background and Related Work}\label{sec:related}

\paragraph{Market impact and execution.}
Almgren--Chriss decomposes impact into permanent and temporary components with linear coefficients fitted to data~\cite{almgren2001optimal}. Obizhaeva--Wang and the propagator literature model impact as a convolution of past order flow with a decaying kernel~\cite{obizhaeva2013optimal,bouchaud2009markets}. Empirically, metaorder impact follows an approximate square-root law with slow decay~\cite{bouchaud2009markets,bacry2015market,zarinelli2015beyond}, kernel shapes are constrained by no-dynamic-arbitrage~\cite{gatheral2010no}, and the functional form is mechanism-dependent, zero then linear in equity \emph{auctions}~\cite{salek2023price}, which is why we treat the square-root law as a venue-specific benchmark rather than ground truth. Bonart~\cite{bonart2026diffusive} makes the causal structure explicit: impact is realized minus counterfactual return, square-root or linear depending on informational coupling. Muhle-Karbe et al.~\cite{muhle2026unified} supply a generative counterpart, decomposing order flow into exogenous informational \emph{core} flow and endogenous \emph{reaction} flow. We use Bonart's identity as the estimand and the core/reaction split as the confounder structure.

\paragraph{Causal amortized inference.}
Prior-data fitted networks meta-learn Bayesian inference by training on synthetic data drawn from a prior over generative processes~\cite{muller2022transformers,hollmann2023tabpfn}. Do-PFN~\cite{robertson2025dopfn} extends this to interventional queries on tabular data, DoT-PFN~\cite{thumm2026causal} to discrete time, and CT-CPFN~\cite{thumm2026towards} to continuous time with irregular schedules.

\paragraph{Microstructure foundation models and market simulators.}
TradeFM \cite{sood2026tradefm} trains a large generative Transformer on billions of trades, reproducing stylized facts~\cite{cont2001empirical} of order flow --- $p(\text{flow} \mid \text{history})$, with no interventional semantics.

\section{Continuous-Time Causal PFN}\label{sec:model}

This section introduces the machinery. The paper's contributions begin with the prior in \S\ref{sec:prior} and the recurrent variant at the end of this section. We build on the continuous-time construction of~\cite{thumm2026towards} and train it on a market-specific prior. We call the resulting trained model \emph{\dopfn{}} (market DoT-PFN), distinguishing it from the general-purpose models it descends from.

\paragraph{Generative process.}
An episode is a linear-drift stochastic differential equation over $d$ variables in topological order,
\begin{equation}\label{eq:ou}
  dX_v = \Big({-\theta_v X_v} + \textstyle\sum_{u \in \mathrm{pa}(v)} w_{vu} X_u\Big)\,dt + \sigma_v\, dW_v,
\end{equation}
where $X_v$ is the state of variable $v$, $\theta_v > 0$ its mean-reversion rate, $\mathrm{pa}(v)$ its parent set with edge weights $w_{vu}$, and the $W_v$ are independent standard Wiener processes with diffusion coefficients $\sigma_v$. Episodes are observed at an irregular schedule $t_1 < \dots < t_T$ drawn from a schedule family (regular, jittered, exponential), integrated by fine-grid Euler--Maruyama~\cite{kloeden1992numerical}. A schedule-invariance criterion~\cite{thumm2026towards} --- predictions depend on the continuous-time process, not the discretization --- allows training on one schedule family and deploying on market bar data.

\paragraph{Interventions.}
An intervention is a first-class input: a target variable $A$, an active window $[\tau_0, \tau_1)$, a kind, and a value. Hard interventions clamp, $\dodo{A(t) := c}$, and \emph{soft} interventions add a constant $\delta$ to the target's drift during the window, the natural formalization of a metaorder, which adds rather than pins flow. Counterfactual training pairs reuse the same Wiener noise with and without the intervention, so the target $Y(t_q)$ under $\dodo{\cdot}$ is exact~\cite{thumm2026interventional}. Post-onset noise is independent of the pre-onset context, so training on such pairs estimates the interventional estimand.

\begin{figure}[t]
  \centering
  \includegraphics[width=0.9\columnwidth]{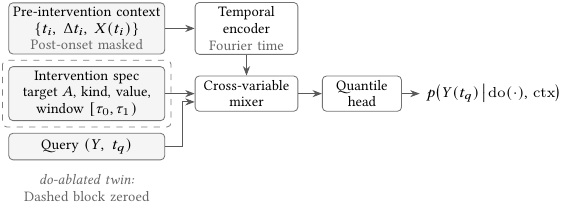}
  \Description{Block diagram of the transformer instantiation. Pre-intervention context tokens enter a temporal encoder with Fourier time features. The intervention specification (target, kind, value, window) and the query (outcome variable and time) enter a cross-variable mixer, and a quantile head outputs the interventional distribution of the queried outcome. The intervention block is dashed to mark the inputs that the do-ablated twin zeroes.}
  \caption{The \emph{transformer} instantiation ($X(t_i)$: observable variables; core $U$ hidden): intervention specification and query enter a cross-variable mixer whose quantile head emits the interventional distribution. The do-ablated twin zeroes the dashed block and nothing else. The recurrent variant (below) replaces the encoder and mixer with $\Delta t_i$-gated recurrences that inject the intervention at each active step.}
  \label{fig:arch}
\end{figure}

\paragraph{Architecture and training (Fig.~\ref{fig:arch}).}
Pre-intervention observations are encoded as tokens carrying the state vector and a Fourier embedding of absolute and inter-observation time, and post-onset observations ($t \ge \tau_0$) are masked. The intervention specification (target, kind, value, window) enters a cross-variable mixer as a distinguished channel. A query names an outcome variable and time. A quantile head is trained with pinball loss on $Y(t_q)$ from the counterfactual pair (4 encoder layers, 4 heads, one cross-variable mixer layer, quantile levels $\{0.1, 0.25, 0.5, 0.75, 0.9\}$, 4.2M parameters; training episodes have $T{=}120$ observations). 

The model therefore amortizes the posterior predictive interventional distribution $p(Y(t_q) \mid \dodo{\cdot}, \text{context})$, marginalizing over the latent Structural Causal Model (SCM) given the observed context. This is the object that ordinary regression cannot recover under hidden confounding (\S\ref{sec:prior-core}).

\paragraph{Recurrent state-space variant.}
The transformer answers a query in one shot: a single attention pass maps context and intervention specification to $Y(t_q)$, and does not explicitly propagate an intervention through a recurrent state --- and on the metaorder benchmark of \S\ref{sec:crypto} its predicted response is indeed nearly flat in execution duration while measured impact accrues. We therefore additionally train a \emph{recurrent state-space} variant: stacked gated delta-product recurrences (the linear-RNN backbone of TempoPFN~\cite{moroshan2025tempopfn}) whose state-transition gates condition on the inter-observation interval $\Delta t_i$, with the intervention injected at every step it is active rather than encoded once (1.94M parameters, same do-ablated-twin protocol). Impact composition then lives in the recurrent state. Its training recipe adds to the prior of \S\ref{sec:prior} a 50/50 soft-intervention mix (\S\ref{sec:controls}), intervention windows of 2 to 30\% of the episode (transformer: 10 to 30\%), and permanent price impact ($\theta_Y{=}0$) in 30\% of episodes. We do not claim the schedule-invariance criterion above for this variant. Its deployment schedules (regular bars) lie inside its training family, and zero-shot robustness beyond them is not established for the checkpoint used here (\S\ref{sec:discussion}). A replicate trained on a mixture of regular, jittered, and exponential schedules recovers it (exponential-schedule loss degradation $+7.8\%$ to $+0.7\%$ at unchanged regular-schedule loss, paired benchmark-correlation cost $-0.02$ $[-0.04, -0.01]$).

\paragraph{Two instantiations.}
The paper carries two matched causal--ablated pairs, each trained once: the transformer, which produces every table row and figure unless explicitly marked otherwise, and the recurrent variant, which contributes only the results labeled as such (the metaorder-benchmark chain of \S\ref{sec:crypto} and the marked rows and sentences of \S\ref{sec:calibration}). The synthetic-tier seed-replication check covers the transformer. The recurrent variant's results quote three-seed ranges in-line. On the synthetic grid the recurrent pair's causal capability likewise sits entirely in the do-pathway (its ablated twin sits within $0.013$ of the pre-window mean at every $\gamma$), but deconfounding is partial: bias $+0.16$ $[0.09, 0.23]$ at $\gamma{=}0$ rising to $+0.52$ $[0.38, 0.66]$ at $\gamma{=}2$, against the headline transformer's $+0.16$ $[0.05, 0.28]$. The two pairs differ in training recipe as well as architecture: a one-shot transformer retrained with the recurrent pair's recipe reaches $+0.80$ $[0.67, 0.93]$ at $\gamma{=}2$ on the same episodes, and the permanent-impact episodes alone move the soft-mix transformer to $+1.04$ $[0.90, 1.18]$, so the gap traces to the recipe that repairs duration accrual, and the recurrent backbone is the least affected by it. The transformer remains the synthetic-tier headline model, and the variant's contribution is the duration-shape axis.

\section{A Market Prior with Controllable Confounding}\label{sec:prior}

This section describes the contribution's prior side: a minimal market whose hidden-information confounding strength is an exactly samplable quantity (\S\ref{sec:prior-core}), and the training distribution built from it (\S\ref{sec:prior-sampling}).

\subsection{Core/reaction flow and informational coupling}\label{sec:prior-core}

The prior's episode is the minimal market that makes impact a genuine causal problem: hidden informational flow $U$, observable order flow $A$, and price $Y$,
\begin{align}
  dU &= -\theta_U\, U\, dt + \sigma_U\, dW_U, \label{eq:core}\\
  dA &= \big({-\theta_A A} + w\,U\big)\,dt + \sigma_A\, dW_A, \label{eq:flow}\\
  dY &= \big({-\theta_Y Y} + \lambda A + \gamma U\big)\,dt + \sigma_Y\, dW_Y. \label{eq:price}
\end{align}
$U$ is the \emph{core} flow of~\cite{muhle2026unified}: exogenous, informational, and hidden --- simulated in the dynamics but masked from the model's tokens. Its two exits are the confounding structure: it drives observable flow ($U \to A$, informed participants trade) and it moves the price directly ($U \to Y$ with weight $\gamma$, information gets into prices whether or not it trades). The flow's own mean reversion and noise ($\theta_A, \sigma_A$) are the endogenous \emph{reaction} component. $\lambda$ is the mechanical impact --- the true causal effect a trader needs.

\paragraph{Calibrated confounding.}
We parameterize the edge weight $w$ by the \emph{core share} $s$: the fraction of $A$'s stationary variance driven by $U$. Solving the stationary Lyapunov equations of \eqref{eq:core}--\eqref{eq:flow} gives the unique $w \ge 0$ realizing any $s \in [0,1)$,
\begin{equation}\label{eq:calibration}
  w \;=\; \sqrt{\frac{s}{1-s}\cdot\frac{\sigma_A^2/(2\theta_A)}{\mathrm{Var}(U)\big/\big(\theta_A(\theta_A+\theta_U)\big)}},
\end{equation}
with $\mathrm{Var}(U) = \sigma_U^2/(2\theta_U)$. Confounding strength is thus an exact, per-episode-samplable quantity. $s=0$ reduces the flow to a plain unconfounded Ornstein-Uhlenbeck (OU) process~\cite{uhlenbeck1930brownian}.

\paragraph{The naive estimator's bias in closed form.}
The observational regression a classical impact model runs --- the stationary Ordinary Least Squares (OLS) slope of $Y$ on $A$ --- is available analytically from the three-variable Lyapunov system~\cite{bhatia2002stability}. Its \emph{confounding component}, the excess over the $\gamma = 0$ slope, is
\begin{equation}\label{eq:bias}
  \mathrm{bias}_{\mathrm{OLS}}(\gamma) \;=\; \frac{\gamma\,\big(\mathrm{Cov}(A,U) + w\,\mathrm{Var}(U)/(\theta_Y + \theta_U)\big)}{(\theta_Y + \theta_A)\,\mathrm{Var}(A)},
\end{equation}
exactly linear in $\gamma$ and exactly zero when $s = 0$. Equation~\eqref{eq:bias} is the law Fig.~\ref{fig:slopes} verifies and the reason this prior needs a PFN: at $\gamma \equiv 0$ the informational-confounding component vanishes (although the filtering offset remains), whereas sampling $\gamma$, $s$, and $\lambda$ per episode forces the model to infer in context where on the confounded--unconfounded spectrum the current market sits.

\paragraph{What the prior identifies.}
With $U$ hidden and no instrument, $E[Y \mid \dodo{A}]$ is \emph{not identified} from observational data, because nothing observable stands in for $U$. The PFN's side information is the prior. It performs amortized Bayesian inference over this known model class, and the prior supplies exactly what the data cannot~\cite{muller2022transformers}. Every claim below should be read in that sense. Matching the prior to the deployment (\S\ref{sec:transfer}) is the mechanism itself.

\subsection{Episode generation and training distribution}\label{sec:prior-sampling}

Episodes sample $\gamma \sim \mathcal{U}(0,2)$, $s \sim \mathcal{U}(0.3, 0.9)$, $\lambda \sim \mathcal{U}(0.25, 1)$, and OU parameters from the ranges of~\cite{thumm2026towards}. One training-distribution choice proved decisive and is reported as a result (\S\ref{sec:synthetic}, \S\ref{sec:transfer}): skewing the $\gamma$ draw toward high confounding ($\gamma = 2\xi^{1/3}$, $\xi \sim \mathcal{U}(0,1)$). Training interventions are hard clamps. The soft (additive-drift) kind enters at evaluation as the metaorder \emph{encoding} (\S\ref{sec:abides}).

\section{Synthetic Proof of Concept}\label{sec:synthetic}

\begin{figure}[t]
  \centering
  \includegraphics[width=0.85\columnwidth]{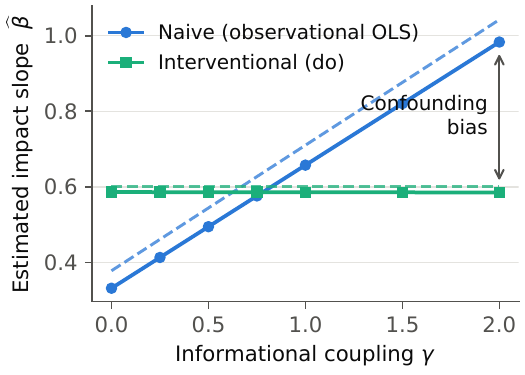}
  \Description{Line chart of the estimated impact slope against the informational coupling gamma from 0 to 2. The naive observational slope rises linearly from about 0.33 to about 0.98 while the interventional slope stays flat near 0.59. Dashed lines show the analytic closed forms, which the measured curves track, and an arrow marks the confounding bias at gamma equal to 2.}
  \caption{The confounding law, measured: naive observational OLS slope (blue) and interventional slope (green) vs.\ informational coupling $\gamma$. Dashed lines are the closed forms (measured $\gamma$-gradient $0.326$ vs.\ $0.332$ analytic, Eq.~\eqref{eq:bias}). Core share $s{=}0.7$; native $Y$-per-$A$ units; per-point SEs $\le 0.011$/$\le 0.023$ ($n{=}2000$/$200$). Transformer pair.}
  \label{fig:slopes}
\end{figure}

\begin{figure*}[t]
  \centering
  \includegraphics[width=0.92\textwidth]{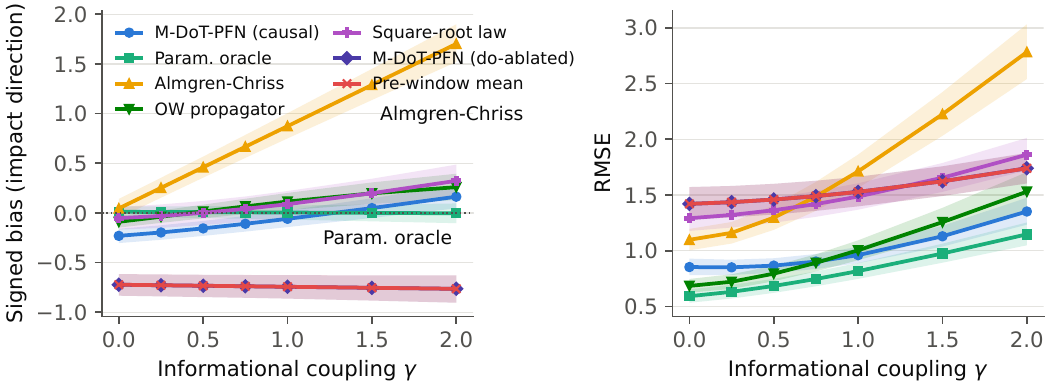}
  \Description{Two line charts against the informational coupling gamma from 0 to 2 with shaded bootstrap bands. Left, signed bias: Almgren-Chriss rises linearly to about 1.7, the square-root law and the OW propagator drift to about 0.3, the causal M-DoT-PFN stays within 0.25 of the unbiased parameter oracle at zero, and the do-ablated twin coincides with the pre-window mean near minus 0.75. Right, RMSE: the oracle is lowest, the causal M-DoT-PFN is next, and it stays below the classical baselines for gamma at or above 1.}
  \caption{Methods on the confounded synthetic market ($n{=}500$ episodes per $\gamma$, identical across methods; bands are episode-bootstrap 95\% CIs; native prior units; transformer pair). Left: signed bias in the impact direction. Right: RMSE against the exact counterfactual.}
  \label{fig:methods}
\end{figure*}

\paragraph{Protocol.}
We train the continuous-time causal PFN (4.2M parameters, 5{,}000 steps, single A100 GPU, $\sim$2h) on the \S\ref{sec:prior} prior, plus an \emph{ablated twin}: the identical architecture, training prior, and training seed with the five do-information inputs (target, kind, value, window ends) zeroed throughout training and prediction --- a pure forecaster that knows where its context ends but not what happens there.

All headline results (Fig.~\ref{fig:methods}, Tables~\ref{tab:transfer} and~\ref{tab:calibration}, except the marked recurrent-variant row of the latter) use one model: the skewed-$\gamma$, hard-intervention configuration of \S\ref{sec:prior-sampling}. Evaluation draws suites of 500 episodes at each pinned $\gamma \in \{0, 0.25, \dots, 2\}$ with $s = 0.7$. All methods predict the same episodes' exact counterfactual outcomes in native prior units, and the transfer tiers of \S\ref{sec:transfer} are standardized to pre-onset volatility units, written $\hat{\sigma}$.

\paragraph{Baselines.} The \emph{true-parameter oracle} is a conditional-mean rollout from the last pre-onset state given the episode's true $(\lambda, \theta_U, \theta_A, \theta_Y)$, with the hidden core at its stationary mean. It is unbiased for the interventional mean but not the minimum-RMSE Bayes predictor, so its RMSE is a conservative reference, not an exact floor.
Further baselines include Almgren--Chriss as the levels-regression permanent-impact estimator; an Obizhaeva--Wang-style (OW) fitted resilient propagator; a \emph{return-space} regression of the horizon-matched return on total signed flow (removing the levels regression's filtering offset); the \emph{square-root law} $\Delta Y = 0.7\,\sigma_{\mathrm{pre}} \sqrt{|Q|/V}$~\cite{bouchaud2009markets} (order size $Q$, volume $V$, pre-onset volatility $\sigma_{\mathrm{pre}}$, written $\hat{\sigma}$ elsewhere) with zero fitted parameters; and last-value (AR1) and pre-window-mean predictors. Where fitting is required, classical baselines are fitted per episode on the pre-onset window only; the square-root reference has no fitted parameters. Nothing is fitted on post-onset outcomes. This family spans execution-cost models, the standard estimation recipe, and the empirical benchmark, so the bias is a property of observational fitting itself, not of one model class.

\paragraph{The law holds (Fig.~\ref{fig:slopes}).}
Two distinct effects act in opposite directions here: a mechanical low-pass \emph{filtering offset} that understates impact even without confounding, and the confounding channel that overstates it as $\gamma$ grows. Our headline bias refers to the confounding channel. The measured naive slope grows linearly in $\gamma$ with slope within 2\% of Eq.~\eqref{eq:bias}, while the measured interventional slope is flat. At $\gamma = 2$ the naive slope exceeds the interventional one by ${\sim}68\%$ (all percentages relative to the interventional slope). Eq.~\eqref{eq:bias} is the $\gamma$-\emph{gradient} of the naive slope, and it is the gradient that the figure confirms. The filtering offset in fact works \emph{against} it: at $\gamma = 0$ the naive slope sits ${\sim}43\%$ \emph{below} the interventional one ($Y$ low-pass-filters fluctuating $A$; a sustained clamp relaxes to the larger steady-state gain $\lambda/\theta_Y$), so across the grid the confounding channel alone swings the naive estimate from a $43\%$ underestimate to a $68\%$ overestimate.

\paragraph{The PFN deconfounds implicitly (Fig.~\ref{fig:methods}).}
The oracle has negligible measured bias at every $\gamma$ ($|\text{bias}| \le 0.013$). The causal \dopfn{} tracks it: bias between $-0.23$ and $+0.16$ across the grid, and the best feasible method on both bias and RMSE for $\gamma \ge 1$ --- the high-confounding regime the prior is built around. At low $\gamma$ the fitted propagator and the unfitted square-root reference are competitive or better. The PFN's advantage is that it does not degrade as confounding opens (square-root bias $+0.32$ at $\gamma{=}2$). The matched-horizon return regression pays a large variance cost (RMSE $2.8$--$5.0$, off the figure's scale). Almgren--Chriss's bias is linear in $\gamma$ (it \emph{is} the naive slope estimator). The propagator is better calibrated but still drifts ($+0.26$ at $\gamma{=}2$). The ablated twin is statistically indistinguishable from the pre-window mean at every $\gamma$ (the mean of the paired per-episode differences is below $10^{-3}$ in magnitude, with bootstrap 95\% CIs within $\pm 10^{-3}$). In this matched ablation, the learned intervention-dependent response resides in the do-pathway. The patterns are robust to training randomness: two additional initialization seeds shift the causal model's bias by at most $0.09$ (RMSE $0.03$) at any $\gamma$, preserve the gradient and its sign change, and move the ablated twin by less than $0.01$.

\subsection{Robustness and controls}\label{sec:controls}

\paragraph{What moved the residual bias --- and what did not.}
Two negative results shape the recipe: training $4\times$ longer changed nothing ($0.396 \to 0.393$ at $\gamma{=}2$), while shifting the training prior's $\gamma$-mass toward high confounding cut bias by 57\% ($0.396 \to 0.169$). This points to the training \emph{distribution} rather than additional training steps as the dominant factor in this comparison. A matched training-side ablation completes the picture: retraining the identical configuration with soft interventions genuinely mixed 50/50 into the training prior roughly halves the unskewed configuration's bias ($0.396$ [$0.27, 0.52$] $\to$ $0.160$--$0.191$ at $\gamma{=}2$ across three training seeds). But it collapses the zero-shot BTC do-head response of \S\ref{sec:transfer} to a seed-dependent $+0.02$--$+0.72\,\hat{\sigma}$ (hard-trained: $+1.42$): the soft-trained response calibrates to the \emph{prior's} impact scale, which under-scales real markets, and semantics-matched training does not yield reliable transfer in this one-shot architecture. A magnitude sweep over all three soft seeds sharpens the reading: rescaling the encoded intervention $0.25\times$--$4\times$ on 173 burst episodes, every seed's response rises monotonically from $0.5\times$ and is significantly nonzero by $2\times$--$4\times$, with attenuation relative to the hard-trained response shrinking as magnitude grows ($1.3$--$1.6\times$ at $4\times$ on two seeds, ${\sim}10\times$ on the weakest). Soft training recalibrates the gain toward the prior's smaller $\lambda$-scale, and the gain is the seed-variable quantity, so the paradox is one of readout rather than a lost pathway. In the recurrent variant of \S\ref{sec:model}, the same semantics-matched recipe \emph{helps} on the metaorder benchmark (curve correlation $+0.94 \to +0.96$; \S\ref{sec:crypto}). The one-shot transformer's transfer evidence below thus rests on encoding-level generalization, not on semantics-matched training.

\paragraph{Off-prior stress test.}
The evaluation above is on-prior by construction, so we generate episodes from three generator families \emph{outside} the training prior's model class --- cyclic price$\to$flow feedback $\mathrm{d}A \mathrel{+}= \kappa Y\,\mathrm{d}t$ ($\kappa \in \{0.1, 0.2\}$), compound-Poisson jumps in the hidden core, and saturating impact ($\lambda A \to 2\lambda \tanh(A/2)$) --- with exact shared-noise counterfactuals at $\gamma{=}1$. An on-prior control through the same standalone simulator reproduces the grid's causal-model bias ($-0.05$ vs.\ $-0.06$), so shifts are attributable to the misspecification itself. Across the four off-prior generators the PFN's bias moves by at most $0.16$, roughly five-fold below the confounding bias being corrected (Almgren--Chriss: $+0.82$--$+0.95$ on the same suites), with the do-head effect intact throughout. Only saturation's CI excludes zero, the expected mild overshoot of a linear-response prior. Table~\ref{tab:stress} lists the per-generator numbers.

\begin{table}[t]
  \caption{Off-prior stress test: causal \dopfn{} bias and paired do-head effect per generator (native prior units, episode-bootstrap 95\% CIs, $n{=}500$ each). Transformer pair.}
  \label{tab:stress}
  \centering
  \begin{tabular}{lcc}
\toprule
Generator & \dopfn{} bias & Do-head effect \\
\midrule
Control (on-prior)      & $-0.05$ $[-0.13, 0.03]$ & $+0.60$ $[0.52, 0.68]$ \\
Feedback $\kappa{=}0.1$ & $-0.02$ $[-0.10, 0.06]$ & $+0.64$ $[0.56, 0.72]$ \\
Feedback $\kappa{=}0.2$ & $+0.03$ $[-0.06, 0.12]$ & $+0.69$ $[0.60, 0.79]$ \\
Core jumps              & $+0.08$ $[-0.02, 0.18]$ & $+0.73$ $[0.64, 0.83]$ \\
Saturated impact        & $+0.11$ $[0.04, 0.18]$  & $+0.57$ $[0.50, 0.64]$ \\
\bottomrule
\end{tabular}
\end{table}

\paragraph{Is the market prior load-bearing?}
A first control is the same paradigm without the market prior: the public general-purpose checkpoints of the discrete-time DoT-PFN~\cite{thumm2026causal}, evaluated zero-shot on identical episodes. On the synthetic grid both sit at bias $-0.81$ to $-0.98$, against the market model's $-0.23$ to $+0.16$. On BTC bursts they collapse onto the ablated reference ($-2.20$ and $-2.22\,\hat{\sigma}$ vs.\ the ablated twin's $-2.26$; market model $-0.84$). A second control isolates the prior exactly: the identical architecture, recipe, and seed retrained on the generic (non-market) prior of~\cite{thumm2026towards}. With its intervention inputs fully intact, it lands on the ablated reference in all three tiers --- synthetic bias $-0.72$ to $-0.76$ (ablated: $-0.72$ to $-0.76$), BTC bursts $-2.26\,\hat{\sigma}$ (ablated: $-2.26$), held-out ABIDES $-0.29\,\hat{\sigma}$, 95\% CI $[-0.64, 0.05]$, evaluated once (ablated: $-0.30$) --- while the market-prior model responds in each. The market prior, with its confounding mass (recipe ablation above) and intervention semantics (encoding ablation of \S\ref{sec:abides}), is what is load-bearing.

\section{Zero-Shot Transfer}\label{sec:transfer}

We now carry the trained model, unchanged, to an agent-based market with exact counterfactual ground truth (\S\ref{sec:abides}), to real market data (\S\ref{sec:crypto}), and to interval calibration (\S\ref{sec:calibration}), with Table~\ref{tab:transfer} as the summary.

\begin{table*}[t]
  \caption{Zero-shot transfer: paired do-head effect (causal \dopfn{} minus do-ablated twin on identical episodes, signed toward the intervention; bootstrap 95\% CIs, $B{=}10^4$; $\hat{\sigma}$ units; $n{=}1000$, $763$, $758$, $757$, and $3{,}802$ episodes per column; Nasdaq column: 50 names, June to August 2026, added for the camera-ready with the same checkpoint, pipeline, and baselines). Baseline deltas are contrasts with the same ablated reference, not causal estimates (protocol: \S\ref{sec:synthetic}--\S\ref{sec:abides}; transformer pair).}
  \label{tab:transfer}
  \centering
  \begin{tabular}{lccccc}
    \toprule
    & ABIDES & BTC bursts & ETH bursts & SOL bursts & Nasdaq bursts \\
    \midrule
    \dopfn{} (causal)      & $\mathbf{+0.38}$ {\scriptsize$[0.34, 0.42]$} & $\mathbf{+1.42}$ {\scriptsize$[1.29, 1.56]$} & $\mathbf{+1.53}$ {\scriptsize$[1.40, 1.67]$} & $\mathbf{+1.22}$ {\scriptsize$[1.10, 1.34]$} & $\mathbf{+1.14}$ {\scriptsize$[1.09, 1.19]$} \\
    OW propagator          & $+0.38$ {\scriptsize$[0.23, 0.53]$} & $+0.88$ {\scriptsize$[0.77, 1.00]$} & $+0.96$ {\scriptsize$[0.86, 1.06]$} & $+0.74$ {\scriptsize$[0.62, 0.85]$} & $+0.18$ {\scriptsize$[0.14, 0.22]$} \\
    Square-root law        & $+0.42$ {\scriptsize$[0.34, 0.51]$} & $+1.86$ {\scriptsize$[1.73, 1.98]$} & $+2.11$ {\scriptsize$[1.99, 2.22]$} & $+1.60$ {\scriptsize$[1.47, 1.73]$} & $+0.76$ {\scriptsize$[0.71, 0.81]$} \\
    Return regression      & $+5.97$ {\scriptsize$[5.49, 6.44]$} & $+3.11$ {\scriptsize$[2.88, 3.34]$} & $+3.27$ {\scriptsize$[3.07, 3.48]$} & $+2.49$ {\scriptsize$[2.31, 2.66]$} & $+0.94$ {\scriptsize$[0.87, 1.00]$} \\
    Almgren--Chriss        & $-0.02$ {\scriptsize$[-0.03, -0.01]$} & $+1.09$ {\scriptsize$[0.98, 1.20]$} & $+1.11$ {\scriptsize$[1.02, 1.20]$} & $+0.89$ {\scriptsize$[0.82, 0.96]$} & $+0.36$ {\scriptsize$[0.34, 0.38]$} \\
    AR1 (last value)   & $+0.05$ {\scriptsize$[-0.03, 0.14]$} & $+1.21$ {\scriptsize$[1.10, 1.33]$} & $+1.41$ {\scriptsize$[1.30, 1.52]$} & $+1.00$ {\scriptsize$[0.88, 1.13]$} & $+0.25$ {\scriptsize$[0.20, 0.30]$} \\
    Pre-window mean    & $-0.00$ {\scriptsize$[-0.00, -0.00]$} & $-0.00$ {\scriptsize$[-0.00, 0.00]$} & $-0.00$ {\scriptsize$[-0.00, 0.00]$} & $-0.00$ {\scriptsize$[-0.00, -0.00]$} & $+0.00$ {\scriptsize$[0.00, 0.00]$} \\
    \bottomrule
  \end{tabular}
\end{table*}

\begin{table}[t]
  \caption{Nominal 80\% interval $[q_{0.1}, q_{0.9}]$: coverage and median width, causal vs.\ do-ablated twin (native units for synthetic rows, $\hat{\sigma}$ for transfer). Binomial noise ${\approx}\pm 0.05$ at $n{=}245$--$500$. ABIDES rows: exploratory seeds (held-out range reserved for Table~\ref{tab:transfer}); crypto rows: June-2026 subset ($n{=}252/245$, SOL not included); Nasdaq row: the full suite of Table~\ref{tab:transfer} ($n{=}3{,}802$). Transformer pair except the marked recurrent row.}
  \label{tab:calibration}
  \centering
  \setlength{\tabcolsep}{5pt}%
  \begin{tabular}{lcccc}
    \toprule
    & \multicolumn{2}{c}{Coverage (nom.\ 0.80)} & \multicolumn{2}{c}{Median width} \\
    \cmidrule(lr){2-3}\cmidrule(lr){4-5}
    & Causal & Ablated & Causal & Ablated \\
    \midrule
    Synthetic $\gamma{=}0$ & 0.77 & 0.76 & 1.68 & 2.44 \\
    Synthetic $\gamma{=}1$ & 0.82 & 0.80 & 2.32 & 3.47 \\
    Synthetic $\gamma{=}2$ & 0.79 & 0.85 & 3.17 & 4.75 \\
    ABIDES                 & 0.32 & 0.39 & 3.20 & 3.73 \\
    ABIDES (recurrent)     & 0.49 & 0.76 & 10.49 & 10.50 \\
    BTC bursts             & 0.47 & 0.35 & 3.06 & 3.46 \\
    ETH bursts             & 0.51 & 0.32 & 3.05 & 3.44 \\
    Nasdaq bursts          & 0.47 & 0.56 & 2.98 & 3.40 \\
    \bottomrule
  \end{tabular}
\end{table}

\subsection{ABIDES: counterfactual ground truth in an agent-based market}\label{sec:abides}

ABIDES~\cite{byrd2020abides} provides what real data cannot: an exact counterfactual. We inject a metaorder agent (fixed-clock market orders, ${\sim}25$--$40\%$ participation over a 15-minute window) into a background market of noise, value, momentum, and adaptive market-making agents, and replay the \emph{identical seed} without its orders. A silent twin of the metaorder agent keeps per-agent seeding and latency identical in the observational run. Pre-onset paths are bit-identical (asserted in automated tests) and post-onset divergence is the causal effect. Episodes are standardized to pre-onset-$\hat{\sigma}$ units. The intervention value is the metaorder's standardized flow rate over the window. Each episode queries the price at one uniformly drawn post-onset bar (30\,s bars, so horizons range from onset to session end) and 1,000 episodes across seeds.

Because the two mechanism findings below (unit standardization, intervention semantics) were discovered by iterating on this environment, all exploratory analysis used seeds $0$--$999$. The numbers here are a single frozen evaluation on the disjoint held-out range $1000$--$1999$. Absolute biases are noise-dominated at this scale, so Table~\ref{tab:transfer} reports the \emph{paired} do-head effect on identical episodes, which cancels shared market noise. The measured seed-replay ground truth on the held-out episodes is a mean impact of $+0.28\,\hat{\sigma}$ (95\% CI $[0.05, 0.50]$; the wide interval reflects seed-to-seed variability of the average impact itself). The causal \dopfn{} responds at $+0.379\,\hat{\sigma}$, 95\% CI $[0.342, 0.416]$, with bias $+0.08\,\hat{\sigma}$ (95\% CI $[-0.27, 0.42]$, containing zero). Bias here is the episode-paired mean error, which is why it is not the difference of the two preceding means. Almgren--Chriss and the naive predictors sit on the ablated reference (deltas $-0.02$ to $+0.05$). The fitted propagator responds at the same magnitude ($+0.38$) but with a ${\sim}4\times$ wider interval ($[0.23, 0.53]$). The square-root law --- ABIDES metaorders at $25$--$40\%$ participation are its home regime --- also responds correctly ($+0.42$, $[0.34, 0.51]$; bias $+0.13$), while the zero-shot return regression overshoots (signed bias $+5.7\,\hat{\sigma}$; Table~\ref{tab:transfer}'s $+5.97$ is its paired delta): its fitted per-flow response does not transfer across scales. The PFN thus matches the best classical mean responses without hard-coding the square-root functional form. The narrow interval on its paired do-head contrast measures precision of the model comparison, not precision of recovered per-episode counterfactual impact.

The paired responses are not a constant offset (all analyses are post-hoc summaries of the single held-out evaluation). Seed-replay noise (per-episode SD $3.6\,\hat{\sigma}$) caps any attainable correlation near $0.1$--$0.2$, and the observed correlations with intervention size ($r{=}0.09$) and realized impact ($r{=}0.11$) sit at that ceiling. The response is thus context-dependent (mean $+0.38$, SD $0.60\,\hat{\sigma}$, strongly bimodal) and averages to the right scale, but the noise floor cannot certify per-episode accuracy beyond that.

Two mechanism findings were necessary conditions. \emph{Units}: raw Agent-based Model (ABM) intervention values reached the model ${\sim}19\,\hat{\sigma}$ out of distribution; standardizing episodes into prior-scale units removed the resulting gross overshoot. \emph{Semantics}: a metaorder \emph{adds} flow (a soft intervention); encoding the same event as a hard clamp cuts the do-head response to $+0.10\,\hat{\sigma}$ (95\% CI $[0.06, 0.15]$; $n{=}300$ exploratory episodes, so the ${\sim}4\times$ ratio is indicative, not a matched ablation), with bias still $-0.31\,\hat{\sigma}$. Encoding the event with its native semantics is, on this evidence, as important as having a do-operator at all.

\subsection{Real markets: detected flow bursts on crypto perpetuals and Nasdaq equities}\label{sec:crypto}

We evaluate on BTC/ETH/SOL perpetuals, the deepest books with complete, public trade-level data. Real markets provide no counterfactual, so the task becomes factual: predict the realized post-event price given the event's signed size, with the paired ablation as the causal readout. We use the public Binance archive
(signed aggregated trades via the \texttt{isBuyerMaker} flag). The archive's labeled liquidation feed has been removed from the public dumps, so events are \emph{detected}: extreme one-sided flow bursts ($|z| \ge 5$, where $z$ is the bar's signed flow minus the trailing 90-minute mean over the trailing standard deviation --- past data only --- and bursts are non-overlapping), which include liquidation cascades. They are plausibly --- not certifiably --- forced. Each event episode covers 90 pre-event and 30 post-event one-minute bars. The intervention value is the burst's mean signed excess flow per bar (relative to the pre-event mean, in standardized units) over its 3-bar window, and the query is one uniformly drawn post-onset bar. April--June 2026, 91 days per symbol: 763 BTC, 758 ETH, and 757 SOL events.

The pattern replicates across all three assets (Table~\ref{tab:transfer}): the do-head effect is $+1.42\,\hat{\sigma}$ (BTC), $+1.53\,\hat{\sigma}$ (ETH), and $+1.22\,\hat{\sigma}$ (SOL); the causal \dopfn{} is less biased than Almgren--Chriss, the fitted propagator, and last-value prediction on every asset (signed bias $-0.84$, $-1.05$, $-0.76$ vs.\ AR1's $-1.05$, $-1.17$, $-0.98$); and the ablated twin again equals the pre-window mean. The return regression sits closer to zero on ETH/SOL ($+0.69$, $+0.51$) but at the worst on-prior RMSE and the largest ABIDES overshoot. The zero-parameter square-root law sits closer to zero on all three ($-0.41$, $-0.47$, $-0.38$; robust to the literature prefactor range). That empirical robustness made it the industry benchmark~\cite{maitrier2025generating} and it is a bar the zero-shot PFN does not yet clear on real data. Splitting each asset's events at its median pre-onset volatility, the effect is stable across regimes (e.g.\ BTC $+1.49$ / $+1.36\,\hat{\sigma}$ calm/stressed; all six CIs overlap), so it is not an artifact of calm markets. The same pipeline and transformer checkpoint carry to equities in follow-up work: on $3{,}802$ detected bursts across 50 Nasdaq names (June to August 2026, 60-second bars, licensed Databento trades) the paired do-head effect is $+1.14\,\hat{\sigma}$ $[1.09, 1.19]$, with calm and stressed halves at $+1.17$ and $+1.10$ (Table~\ref{tab:transfer}). On this wider panel the model over-reacts (signed bias $+0.48\,\hat{\sigma}$ against the ablated twin's $-0.65$; the ten largest names sit at $+0.01$): the paired gain replicates, the level does not. A sign-shuffle control separates impact from trend-following: randomly flipping event signs sends the measured directional move from $+1.05$ to $-0.13\,\hat{\sigma}$ (impact is direction-locked, not repackaged volatility), leaves the ablated twin unbiased on the shuffled labels, and makes the causal model manufacture error by following the now-wrong labels --- the response tracks the labeled direction, not the realized trend.

As a population-level complement to detected bursts, we reconstruct synthetic metaorders from the public tape (child orders assigned to randomly drawn synthetic traders; a metaorder is a trader's same-sign run~\cite{maitrier2025generating}) and compare the paired do-head response against the \emph{measured} realized impact across $12$ duration\,$\times$\,magnitude cells ($2{,}158$ BTC episodes): the curve correlation is $+0.68$ (episode-bootstrap 95\% CI $[0.60, 0.75]$, $B{=}2000$), yielding a reusable real-data benchmark for the duration-dependent response. Targets are population means of selected events, so this tier shares the burst suite's factual-outcome caveats.

The residual mismatch this benchmark localizes, a duration-flat one-shot response against duration-accruing measured impact, is what the recurrent variant of \S\ref{sec:model} repairs. Its fixed-size response accrues natively, rising $3.1\times$ over durations of 2--31 bars where the benchmark's per-unit-flow measured impact rises ${\sim}3.6\times$. The same 12-cell curve correlation (a correlation of cell means, not an episode-level fit; cells re-formed per bootstrap resample) reaches $+0.94$ hard-trained (episode-bootstrap 95\% CI $[0.90, 0.96]$) and $+0.96$ with the semantics-matched soft mix ($[0.93, 0.98]$; three-seed range $+0.92$--$+0.96$), against the transformer's $+0.68$ $[0.60, 0.75]$ on identical episodes, a paired difference of $+0.28$ $[0.21, 0.34]$ of which per-step intervention injection into an otherwise one-shot transformer supplies $+0.22$ $[0.16, 0.27]$ and recurrence the remaining $+0.06$ $[0.03, 0.10]$. The shape replicates at $+0.97$ $[0.93, 0.98]$ on 40 held-out days (May 11--June 19) that entered no benchmark run during variant development, with no refit of any kind. It also replicates on ten Nasdaq names ($7{,}162$ synthetic metaorders): one-shot $+0.62$ $[0.54, 0.69]$, recurrent $+0.95$ $[0.90, 0.97]$ (paired difference $+0.32$ $[0.27, 0.37]$), with a Nasdaq $k$ of $0.28$. The two simulation tiers with known counterfactuals also show this ordering on analogous 12-cell benchmarks. On $4{,}000$ fresh prior episodes the correlation is $+0.94$ (one-shot) and $+0.98$ (recurrent), close together because the exact effect saturates within a few bars, and the recurrent $k$ is $0.73$ in native units. On $600$ fresh ABIDES seed-replay pairs with execution windows of 3 to 24 bars it is $+0.69$ and $+0.97$, a paired difference of $+0.28$ $[0.09, 0.46]$ (Table~\ref{tab:evidence}).

The recurrent variant's remaining mismatch on the duration benchmark is primarily in level rather than shape; this is not yet zero-shot transfer of absolute causal magnitude. Denote by $k$ the per-domain constant that rescales its response onto the measured scale: fitted on 20 days of BTC episodes, $k$ transfers to 20 disjoint days (calibrated level $1.19\times$, corr $+0.98$) and is near-universal within crypto ($k = 0.49/0.51/0.53$ across venues, ETH/SOL zero-shot). The crypto $k$ is fitted to measured --- observational --- impact, so calibrated levels inherit any confounding in that target. Only the ABIDES $k$ (\S\ref{sec:calibration}) is fitted against counterfactual ground truth. On the burst tier the variant's uncalibrated response reaches the measured scale on two of three training seeds (causal bias $-1.2$ to $+0.6\,\hat{\sigma}$ across seeds and assets, vs.\ the transformer's uniform $-0.8$ to $-1.1$ undershoot; per-seed do-head deltas $+1.2$--$+3.0\,\hat{\sigma}$). Ensembling the three seeds is the strongest honest number on this tier (ensemble bias $+0.02/-0.22/+0.13\,\hat{\sigma}$; BTC/SOL intervals span zero, ETH's upper bound $-0.00$; assets cross-correlated per the multiplicity note below).

A placebo test probes whether the detection pipeline, rather than the events, produces these patterns: we sample time-shifted non-event anchors ($|z| < 2$, well separated from every detected burst; $654$--$663$ per asset) and attach fake interventions resampled from the real events' size distribution (the placebo pipeline re-draws query bars, so its real-event numbers differ slightly from Table~\ref{tab:transfer} and the biases above). At placebo anchors the ablated twin is unbiased ($-0.03\,\hat{\sigma}$: nothing systematic happens at non-events) while the causal model, fed a fake intervention, \emph{creates} $+1.35\,\hat{\sigma}$ of bias. At real events the same response \emph{reduces} bias from $-2.29$ to $-0.95$. ETH/SOL replicate (placebo ablated $-0.10$/$+0.10$, causal placebo $+1.34$/$+1.23$, real-event reduction $-2.59 \to -1.15$, $-1.96 \to -0.83$). The do-head's response is only accurate where a real event occurred.

On raw RMSE the square-root law is best ($2.1$--$2.5\,\hat{\sigma}$) --- expected for a factual task on recency-dominated data. We therefore do not claim superior factual forecasting accuracy: our contribution is a learned intervention-conditioned distribution under a specified structural prior. We state the epistemic status plainly: with no counterfactual on real data, these results establish \emph{predictive bias reduction from the do-information, consistent with, but not proof of, causal transfer}; post-burst continuation driven by momentum cannot be excluded, and we frame the crypto evidence as the paired ablation and bias ranking, not ground-truth causal recovery. On multiplicity: the per-asset deltas replicate one pre-specified comparison across three cross-correlated assets (liquidation cascades are market-wide, so not independent). We report unadjusted intervals, and every headline delta exceeds twenty bootstrap standard errors, so any standard correction leaves the conclusions unchanged. A semi-synthetic check supplies a counterfactual under an imposed impact kernel, not the unobserved natural-market counterfactual: metaorders with a known impact kernel injected into real Nasdaq burst contexts (701 episodes per arm, 24 names) are recovered by a soft-mix-trained transformer seed at correlation $0.54$ to $0.56$, with scale over-predicted $1.6$ to $1.7\times$.

\subsection{Counterfactual calibration}\label{sec:calibration}

Table~\ref{tab:calibration} asks whether the quantile head's counterfactual intervals are honest. In-distribution they are: coverage of the nominal 80\% interval is $0.77$--$0.82$ across the whole confounding range, at roughly one-third narrower intervals than the do-ablated twin. Under zero-shot transfer the picture inverts (causal coverage $0.32$ on ABIDES, $0.47$--$0.51$ on crypto, $0.47$ on Nasdaq; ablated $0.32$--$0.39$, and $0.56$ on Nasdaq, whose wider intervals cover more than the over-reacting causal model): intervention-sensitive point responses transfer, whereas raw interval coverage does not, because the prior's outcome-noise scale is not the ABM's or the market's. The limited benefit of seed ensembling suggests that model-seed variability alone does not explain the coverage gap: a predictive mixture over three training seeds covers $0.35$ (transformer) and $0.50$ (recurrent) on ABIDES against single-seed values of $0.33$--$0.34$ and $0.46$--$0.54$, so ensembling is not the interval repair.
Split-conformal recalibration~\cite{romano2019conformalized} offers a concrete fix: widen both interval ends by an empirical quantile of held-out nonconformity scores $E_i = \max\{q_{0.1}(x_i) - y_i,\; y_i - q_{0.9}(x_i)\}$, requiring exchangeability of calibration and test examples for its marginal coverage guarantee, and leaving point predictions untouched. On ABIDES the calibration set can even be \emph{counterfactual} (seed replay supplies the outcome under $\dodo{\cdot}$): calibrating on seeds $0$--$499$ and evaluating on $500$--$999$ restores nominal-80\% coverage from $0.31$ on that evaluation half to $0.82$, at ${\sim}4\times$ wider intervals. On real markets only factual outcomes exist, so the same recipe can provide marginal factual coverage under exchangeability, while counterfactual validity still rests on the prior.

The recurrent variant's raw ABIDES coverage is higher ($0.49$ vs.\ the transformer's $0.32$) and split-conformal restores $0.81$ at only $1.6\times$ width --- but from a ${\sim}3\times$ wider base, so the transformer's conformalized intervals remain the narrower ones in absolute terms (transformer $13.1$ vs.\ recurrent $16.9\,\hat{\sigma}$), and, unlike in-distribution, the variant's do-pathway \emph{worsens} its raw ABIDES calibration against its own ablated twin ($0.49$ vs.\ $0.76$ at equal width; Table~\ref{tab:calibration}). Its scale constant is likewise per-domain, not global: the crypto $k$ ($0.49$--$0.53$) does not transfer to ABIDES, where $k$ is several times smaller and varies with the training seed ($0.049$--$0.096$) and with the execution protocol ($0.23$ on the duration sweep of \S\ref{sec:crypto}). Fitting $k$ on $n{=}500$ calibration episodes and applying it to $n{=}1000$ held-out seed-replay counterfactuals makes the calibrated bias seed-invariant at $-0.12\,\hat{\sigma}$ with all three seeds' 95\% CIs spanning zero --- though the calibrated response sits at the low end of the measured CI, so the transformer remains the better ABIDES model. Scale and interval calibration compose: recentering by $k$ before conformalizing keeps coverage nominal ($0.783$ at width $13.4$ against $0.812$ at $16.9$ unscaled; $n{=}300$ vs.\ $n{=}500$) on counterfactual seed-replay ground truth. Cross-domain scale inference, predicting $k$ from the observable tape instead of fitting it, remains open. Its labeling cost is small: on a Nasdaq closing-auction tier, five labeled auctions fit a $k$ whose held-out RMSE ($0.240$) beats a last-value predictor ($0.280$).

\begin{table*}[t]
  \caption{What each evaluation tier establishes (numbers as reported in the text; transformer pair unless marked).}
  \label{tab:evidence}
  \centering
  \footnotesize
  \setlength{\tabcolsep}{4pt}%
  \begin{tabular}{p{3.0cm}p{2.8cm}p{3.4cm}p{3.6cm}p{2.8cm}}
    \toprule
    Aspect & Synthetic prior (\S\ref{sec:synthetic}) & ABIDES seed replay (\S\ref{sec:abides}) & Crypto bursts (\S\ref{sec:crypto}) & Nasdaq bursts (\S\ref{sec:crypto}) \\
    \midrule
    Counterfactual truth & Exact (shared noise) & Exact (seed replay) & None (factual) & None (factual) \\
    Deconfounding & Bias $-0.23$ to $+0.16$ & Bias $+0.08$ $[-0.27, 0.42]$ & Not identifiable & Not identifiable \\
    Do-head response & $+0.60$ $[0.52, 0.68]$ & $+0.38$ $[0.34, 0.42]$ & $+1.22$ to $+1.53$ & $+1.14$ $[1.09, 1.19]$ \\
    Duration shape (correlation, one-shot to recurrent) & $+0.94$ to $+0.98$ & $+0.69$ to $+0.97$ & $+0.68$ to $+0.96$ & $+0.62$ to $+0.95$ \\
    Signed bias of the level & Native units & At the measured scale & $-0.8$ to $-1.1$ & $+0.48$ \\
    Scale constant $k$ (recurrent) & $0.73$ (native units) & $0.05$ to $0.23$ & $0.49$ to $0.53$ & $0.28$ \\
    Interval calibration & $0.77$ to $0.82$ & $0.32$, conformal $0.82$ & $0.47$ to $0.51$, factual only & $0.47$, factual only \\
    \bottomrule
  \end{tabular}
\end{table*}

\section{Discussion and Limitations}\label{sec:discussion}

This is a proof of concept for the interventional axis of market foundation models: the prior is a stylized three-variable market with no order-book queueing, cancellations, or inventory, trades-as-drift-shifts idealize execution mechanics, the detected crypto events are only plausibly exogenous, and we do not compete with large observational models~\cite{sood2026tradefm} on scale or realism. Two further limitations: the recurrent variant's schedule robustness outside regular bars requires schedule-mixture training (\S\ref{sec:model}), and, as the placebo's flip side, fed a mislabeled event the do-head confidently manufactures impact, so deployment requires trustworthy event labels. Table~\ref{tab:evidence} separates the transfer questions by tier: whether the model responds to an intervention, whether its response has the correct temporal shape, and whether its absolute magnitude is identified, calibrated, and covered by honest intervals. A positive paired do-head effect across these tiers while the ablated twin collapses onto the mean supports the narrower, load-bearing claim: a PFN can amortize causal inference under hidden confounding, where identification is supplied by the prior, not the data, and carry it zero-shot into markets it never saw, \emph{when the prior's confounding mass and intervention encoding are sufficiently aligned with the deployment event}. The observed duration-shape gap was substantially reduced by injecting the intervention at every active step and further improved by recurrence (\S\ref{sec:crypto}), and the residual level gap is a per-domain scale constant whose in-context inference remains open (\S\ref{sec:calibration}). Within this proof of concept, the transformer pair provides the stronger ABIDES counterfactual-level results, while the recurrent variant better matches duration profiles; neither yet establishes deployable causal execution-cost estimates from observational market data alone. That no single instantiation yet achieves both is the sharpest open problem.

The flip side of prior-based identification cuts both ways. In a companion experiment (identical recipe), \emph{enriching} the prior with jump-diffusion in the hidden core suppressed the zero-shot do-head response roughly three- to eight-fold across the transfer tiers while fitting its own prior well: components that offer alternative explanations for intervention-sized moves dilute the prior-supplied identification, so predictive fit does not certify the causal readout, and prior \emph{match}, not richness, is the criterion. The converse exposure is misspecification: the deconfounding is only as trustworthy as the model class the prior spans, and our prior assumes linear-drift OU dynamics where real microstructure is nonlinear, heavy-tailed, nonstationary, and shot through with feedback from price back into flow~\cite{bouchaud2018trades}. The stress test of \S\ref{sec:controls} and the transfer tiers probe exactly this misspecification: feedback, jumps, and saturation lie outside the prior's class, and neither ABIDES nor crypto is an OU system. The point-level causal response degrades gracefully (bias shifts $\le 0.16$, \S\ref{sec:controls}) while calibrated uncertainty does not.

Excluded from the headline prior by design, pending a causal-attribution ablation for any realism layer: jump-diffusion drift (tested and rejected above), factor-structured multi-asset DAGs, and Hawkes-timed interventions. An ABM analogue of the $\gamma$-sweep and the rough/fractional scaling limit of~\cite{muhle2026unified} are future work.

\paragraph{Reproducibility and data availability.}
All code, checkpoints, and result files are available at \url{https://github.com/thummd/causal-market-simulator}.
The crypto data is the public Binance archive (\url{https://data.binance.vision/}) with dates, symbols, and detector settings in \S\ref{sec:crypto}. The Nasdaq trades are licensed from Databento (\url{https://databento.com/}) and are not redistributed. The pull manifest and the equity scripts are in the repository.

\begin{acks}
Artificial Intelligence (AI) assisted with scripting and drafting. The authors audited, tested, and validated all AI-generated code and text. We would like to thank Ruben Wiedemann and Efstratios Manolakis for inspiring discussions and feedback as well as the anonymous reviewers for their helpful comments.
\end{acks}

\bibliographystyle{ACM-Reference-Format}
\bibliography{refs}

\end{document}